\documentclass[twoside,11pt]{article}

\usepackage{blindtext}

\usepackage[preprint]{jmlr2e}

\usepackage[htt]{hyphenat}
\usepackage{lastpage}
\usepackage{url}
\jmlrheading{23}{2022}{1-\pageref{LastPage}}{1/21; Revised ?}{?}{21-0000}{Picheny, Berkeley, Moss, Stojic, Granta, Artemev, Ober, Ghani, Goodall, Paleyes, Vakili, Pascual-Diaz, Markou, Qing, Loka \& Couckuyt}

\usepackage{subcaption}
\usepackage{tablefootnote}
\ShortHeadings{Trieste: Efficiently Exploring The Depths of Black-box Functions with TensorFlow}{Picheny et al.}
\firstpageno{1}

\begin{document}

\title{Trieste: Efficiently Exploring The Depths of Black-box Functions with TensorFlow}

\author{\name Victor Picheny \email victor@secondmind.ai  \\
       \name Joel Berkeley \\%\email artem@secondmind.ai \\
        \name Henry B. Moss \\%\email henry.moss@secondmind.ai \\
        \name Hrvoje Stojic \\%\email artem@secondmind.ai \\
        \name Uri Granta \\%\email artem@secondmind.ai\\  
       \name Sebastian W. Ober  \\%\email artem@secondmind.ai\\
       \name Artem Artemev \\%\email artem@secondmind.ai  \\
       \name Khurram  Ghani\\%\email artem@secondmind.ai \\
       \name Alexander Goodall\\% \email artem@secondmind.ai \\
        \name  Andrei Paleyes\\%  \email artem@secondmind.ai  \\
       \name Sattar Vakili \\
       \name Sergio Pascual-Diaz\\%\email artem@secondmind.ai \\
       \name Stratis Markou \\
        \addr Secondmind, Cambridge, UK 
        \AND
        \name Jixiang Qing\\%  \email artem@secondmind.ai  \\
        \name Nasrulloh R.B.S Loka \\% \email artem@secondmind.ai  \\
        \name Ivo Couckuyt\thanks{This research receives funding from the Flemish Government under the “Onderzoeksprogramma Artifciele Intelligentie (AI) Vlaanderen” programme.} \\%\email artem@secondmind.ai \\
       \addr Ghent University - imec, Ghent, Belgium%\footnote{Ghent University - imec is partially funded by Flanders AI.}
    }

\editor{?}

\maketitle

\begin{abstract}%   <- trailing '%' for backward compatibility of .sty file
We present \textsc{Trieste}, an open-source Python package for Bayesian optimization and active learning benefiting from the scalability and efficiency of \textsc{TensorFlow}. Our library enables the plug-and-play of popular \textsc{TensorFlow}-based models within sequential decision-making loops, e.g. Gaussian processes from \textsc{GPflow} or \textsc{GPflux}, or neural networks from \textsc{Keras}. This modular mindset is central to the package and extends to our acquisition functions and the internal dynamics of the decision-making loop, both of which can be tailored and extended by researchers or engineers when tackling custom use cases. 
\textsc{Trieste} is a research-friendly and production-ready toolkit backed by a comprehensive test suite, extensive documentation, and available at \url{https://github.com/secondmind-labs/trieste}.
\end{abstract}

\section{Introduction}% 1. why does Trieste exist? (who is this for (TF dev + researchers(e.g. modualtiryt + new models))) + computational efficiency (GPU) 

\textsc{TensorFlow} is one of Python's primary machine learning frameworks, offering both flexibility and scalability through its support for auto-differentiation and GPU-based computation. Yet, \textsc{TensorFlow} 2 does not have a library for Bayesian Optimization (BO) --- an increasingly popular method for black-box optimization under heavily constrained optimization budgets \citep[see][for an introduction]{shahriari2016taking}. This lack of support is likely due to the inherently sequential and evolving nature of active learning loops, which makes BO implementations prone to trigger expensive retracing of the computational graphs, as used by \textsc{TensorFlow} to accelerate numerical calculations. However, once special care is taken to avoid unnecessary retracing, a \textsc{TensorFlow}-based BO library would allow users not only to leverage versatile and powerful (probabilistic) \textsc{TensorFlow} modeling  libraries (e.g. \textsc{Keras}, \textsc{GPflow}, \textsc{GPflux}), but also to benefit from BO-specific perks like the freedom to define acquisition functions without specifying their gradients, easily parallelized optimization of acquisition functions, and in-the-loop monitoring of models and convergence statistics (e.g. \textsc{TensorBoard}).

In this paper, we present \textsc{Trieste}\footnote{Trieste was the first crewed vessel to reach the bottom of the Mariana trench --- the literal global minimum.}, a highly modular, flexible and general-purpose BO library designed to enable users working within \textsc{TensorFlow} ecosystems to 1) deploy their own existing models to drive BO loops and 2) build BO pipelines that can harness the ease and computational efficiency provided by \textsc{TensorFlow}'s automatic differentiation and support for modern compute resources like GPUs. Our library is oriented towards real-world use and contains a wide range of advanced BO functionalities, with a focus on modularity to allow ease of extension with custom models and acquisition functions. %matched only by \textsc{Torch}-based \textsc{BoTorch} of \cite{balandat2020botorch} %(see Section \ref{comparison}) which, although widely regarded as the \textit{state-of-the-art} BO Python library, does not support the large number researchers and engineers with \textsc{TensorFlow} models and pipelines.
% Ultimately, \textsc{Trieste}

\section{Related Work}
\label{sec:related}
Many open-source libraries have been built to support the recent increase in the use and development of BO methodology. For example, Python users with models written in \textsc{Torch} or \textsc{NumPy} can easily find compatible libraries such as \textsc{BoTorch} \citep{balandat2020botorch}, \textsc{GPyOpt} \citep{gpyopt2016}, \textsc{RoBo} \citep{klein2017robo}, \textsc{Emukit} \citep{paleyes2021emulation} or \textsc{Dragonfly} \citep{kandasamy2020tuning}. Similarly, those running \textsc{R}, \textsc{C}\texttt{++} or \textsc{Java}  can use \textsc{DiceOptim} \citep{roustant2012dicekriging}, \textsc{BayesOpt} \citep{martinez2014bayesopt} or \textsc{SMAC} \citep{hutter2011sequential}. 
The library \textsc{GPflowOpt} \citep{knudde2017gpflowopt} was built on \textsc{TensorFlow} 1 with an intent similar to \textsc{Trieste}, but is not actively maintained anymore and does not support the fundamentally different \textsc{TensorFlow} 2.

%Unfortunately, missing from the above list is a BO library that supports \textsc{TensorFlow} 2, one of the most popular frameworks for performing machine learning in Python. 
% Note that \textsc{TensorFlow} 2 differs fundamentally to \textsc{TensorFlow} 1 and so is not supported by \textsc{GPflowOpt}. %Moreover, many of these above libraries are no longer maintained (GPyOpt, GPflowOpt, RoBo), or target only hyper-parameter optimization  (\textsc{Dragonfly} and \textsc{SMAC}).

\section{Key features and Design}% 3. internals (flexibility + moldaulriyty) (rules + builders)

We now present the modular structure of \textsc{Trieste} which contains four key building blocks, a choice of high-level interface (either \texttt{AskTellOptimizer} or \texttt{BayesianOptimizer}), a choice of \texttt{ProbabilisticModel}, and a pairing of \texttt{AcquisitionRule} and \texttt{AcquisitionFunction}. Although \textsc{Trieste}'s structure allows a high level of customization, sensible defaults are provided throughout the library in order to give new users good starting points. 

\subsection{Interfaces for different levels of control over function evaluation}

\begin{figure*}%
\subfloat[\texttt{AskTellOptimizer}]{\includegraphics[height= 0.2\textwidth]{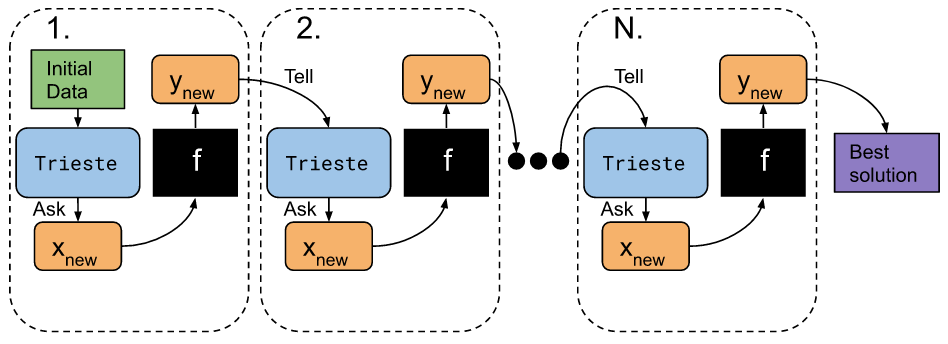}}%
\hspace{10pt}\subfloat[\texttt{BayesianOptimizer}]{\includegraphics[height= 0.2\textwidth]{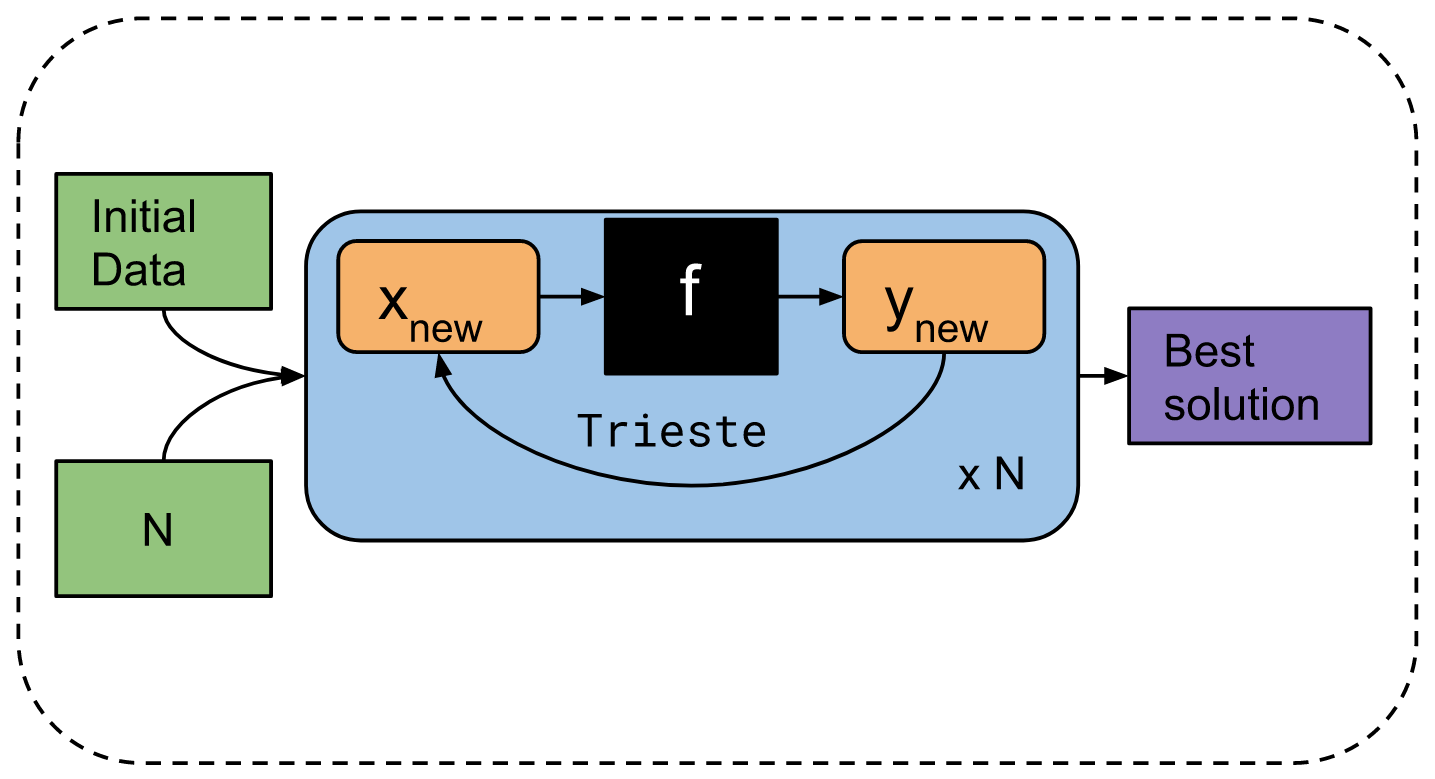}}%
\caption{\textsc{Trieste}'s different interfaces over $N$ optimization steps. The \texttt{AskTellOptimizer} requires users to make manual evaluations, which is useful for real-world settings, e.g. evaluating the objective function requires laboratory or distributed computing resources. In contrast, the \texttt{BayesianOptimizer} queries the black box directly, performing all $N$ BO iterations without user interaction.} %
\label{fig:structure}%
\end{figure*}

A key design choice of \textsc{Trieste} is its \texttt{AskTellOptimizer} interface, which need not have direct access to the objective function. In many libraries, the objective must be a query-able function to be passed into the loop and called for each BO step, an assumption that is rarely suitable when performing BO in the real world. In contrast, an \texttt{AskTellOptimizer} outputs recommended query points (the \textit{ask}) and then waits for the user to return new evaluations (the \textit{tell}) (see Figure~\ref{fig:structure}). This interface allows \textsc{Trieste} users to apply BO across a range of non-standard real-world settings,
e.g., when evaluating the objective function requires laboratory \citep{jordens2023optimization} or distributed computing resources, such that users have only partial control over the environment \citep{qing2022robust} or batches of evaluations arrive asynchronously \citep{kandasamy2018parallelised}. For settings where it is appropriate for the objective function to be passed into the BO, e.g., for experiments with synthetic problems, we also provide a more standard \texttt{BayesianOptimizer} interface that will run multiple BO steps.

\subsection{Versatile Model Support}

The models in the BO loop can be any model written in \textsc{TensorFlow}, and \textsc{Trieste} is designed to make it easy to add a new model, through a set of general model interfaces.
We provide direct interfaces to import models from well-established \textsc{TensorFlow} modeling libraries, e.g., Gaussian processes from \textsc{GPflow} \citep{matthews2017gpflow} and \textsc{GPflux} \citep{dutordoir2021gpflux}, as well as neural networks from \textsc{Keras}. 
\textsc{Trieste} users have a range of popular probabilistic models from these libraries available out of the box.
They cover both regression and classification tasks and range from standard GPs \citep{rasmussen} to alternatives like sparse variational GPs \citep{hensman2014scalable}, Deep GPs \citep{salimbeni2017doubly} or Deep Ensembles \citep{lakshminarayanan2016simple}, that scale much better with the number of function evaluations.
Finally, we provide user-friendly model builders with sensible default setups, allowing users to get more quickly to good results and facilitate usage for those with less experience with probabilistic models.
Hence, \textsc{Trieste} users can benefit from a large choice of models with a wide range of complexity, unlocking novel applications for BO.

Importantly, \textsc{Trieste}'s BO loops allow the modeling of multiple quantities, using either multiple separate models or a single multi-output model. This framework naturally supports common BO extensions like multi-objective optimization \citep{knowles2006parego}, multi-fidelity optimization \citep{swersky2013multi}, optimization with constraints \citep{schonlau1998global}, and combinations thereof.

\subsection{Acquisition Rules and Functions}

Regardless of the interface and model choice, the recommendation of query points is controlled by an \texttt{AcquisitionRule}. 
While the vanilla BO rule is to query the point that optimizes a particular \texttt{AcquisitionFunction}, the \texttt{AcquisitionRule} is a useful abstraction to handle complex cases for which a high level of flexibility is needed. For example, Trieste includes variable optimization spaces for \texttt{AcquisitionFunction} \citep[e.g., using trust regions,][]{diouane2022trego}, a multi-step procedure for selecting query points \citep[][]{binois2021portfolio}, and a greedy approach to build batches of query points \citep{moss2021gibbon}.

A wide range of acquisition functions are already provided in \textsc{Trieste} to tackle most of the usual BO cases, with many based on the gold-standard Expected Improvement \citep[EI; ][]{jones1998efficient}, including variants for batch \citep{chevalier2013fast, gonzalez2016batch, balandat2020botorch}, noisy \citep{huang2006global}, multi-objective \citep{daulton2020differentiable} and constrained \citep{gardner2014bayesian} optimization.
%exact and Monte-Carlo multi-point EI \citep{chevalier2013fast, balandat2020botorch} and local penalization \citep{gonzalez2016batch} for batch BO, augmented EI \citep{huang2006global} for noisy optimization, Monte-Carlo Expected Hyper-volume Improvement \citep{daulton2020differentiable} for (batch) multi-objective BO, and constrained EI \citep{gardner2014bayesian} for constrained optimization.
%To demonstrate the ease of designing new acquisition functions in \textsc{Trieste}, w
We also include implementations of recent information-theoretic approaches for multi-fidelity optimization \citep{moss2021mumbo}, a scalable extension of batched Thompson sampling \citep{vakili2021scalable}, 
as well as popular active learning methods for improved classification \citep{houlsby2011bayesian} or contour line estimation \citep{picheny2010adaptive}.

\textsc{Trieste} is designed to make it straightforward for a user to specify a new acquisition function or rule. 
Automatic differentiation directly provides the function gradients, which are leveraged by \textsc{Trieste}'s supported acquisition function optimizers, including an effective parallelized multi-start \textrm{L-BFGS-B }optimizer. Special care is taken to allow \texttt{AcquisitionFunction}s to be updated without expensive retracing of the computational graphs for each BO step.

\section{Conclusions and Future Plans}

\textsc{Trieste} is an open-source project that allows \textsc{TensorFlow} practitioners to easily use BO in their systems. It is a highly flexible library designed with modularity in mind, easy to extend with custom models and acquisition functions. Backed by continuous integration and comprehensive unit tests ($97\%$ coverage), \textsc{Trieste} is a reliable and robust framework used for both real-world deployment and research  has recently been taken up by researchers to develop new BO methodology \citep{vakili2021scalable, picheny2022bayesian, chang2022fantasizing, heidari2022finding, moss2021gibbon, moss2022information, moss2023inducing, qing2022robust, qing2022spectral, qing2023} whilst also being used across a range of applications including  designing heat exchangers \citep{paleyes2022penalisation} and  improving adhesive bonding \citep{jordens2023optimization}.

\textsc{Trieste} relies on features added and improved by the community and so we gladly welcome feature requests and code contributions. We plan to continue to add new functionality orientated to supporting the application of BO in the real world. In the near term, this will include high-dimensional objective functions \citep{binois2022survey} and non-Euclidean search spaces \citep{moss2020boss,ru2020bayesian,deshwal2021mercer}.

% \section*{Acknowledgments}
% We want to thank Secondmind for supporting the open-sourcing effort.

\bibliography{sample}
\end{document}